\documentclass{article}

\usepackage{microtype}
\usepackage{graphicx}
\usepackage{subcaption}
\usepackage{booktabs} 

\usepackage{hyperref}
\usepackage{multirow}
\usepackage{soul, color, xcolor}
\usepackage{colortbl}
\usepackage[table]{xcolor}
\usepackage{pifont}

\definecolor{lightgray}{gray}{0.9}
\usepackage{array}
\definecolor{lightblue}{RGB}{91,155,213}

\usepackage[preprint]{icml2026}

\usepackage{amsmath}
\usepackage{amssymb}
\usepackage{mathtools}
\usepackage{amsthm}

\usepackage[capitalize,noabbrev]{cleveref}

\theoremstyle{plain}

\theoremstyle{definition}

\theoremstyle{remark}

\usepackage[textsize=tiny]{todonotes}

\begin{document}

\twocolumn[
  \icmltitle{Robust MLLM Unlearning via Visual Knowledge Distillation}



  \icmlsetsymbol{equal}{*}

  \begin{icmlauthorlist}
    \icmlauthor{Yuhang Wang}{Xidian}
    \icmlauthor{Zhenxing Niu}{Xidian}
    \icmlauthor{Haoxuan Ji}{XJTU}
    \icmlauthor{Guangyu He}{Xidian}
    \icmlauthor{Haichang Gao}{Xidian}
    \icmlauthor{Gang Hua}{Amazon}
  \end{icmlauthorlist}

  \icmlaffiliation{Xidian}{Xidian University, China}
  \icmlaffiliation{XJTU}{XJTU University, China}
  \icmlaffiliation{Amazon}{Amazon.com, USA}

  \icmlcorrespondingauthor{Zhenxing Niu}{zxniu@xidian.edu.cn}

  \icmlkeywords{Machine Learning, ICML}

  \vskip 0.3in
]



\printAffiliationsAndNotice{}  

\begin{abstract}

Recently, LLM unlearning approaches have been proposed to remove sensitive information from well-trained large models. However, unlearning for Multimodal Large Language Models (MLLMs) remains at an early stage. Inspired by recent studies on the internal mechanisms of MLLMs, we propose to disentangle visual and textual knowledge within MLLMs and introduce a dedicated approach that selectively erases target visual knowledge while preserving textual knowledge.
Unlike previous unlearning methods that rely on output-level supervision, our approach introduces a \emph{Visual Knowledge Distillation} (VKD) scheme, which leverages intermediate visual representations within the MLLM as supervision signals. This design substantially enhances both unlearning effectiveness and model utility. Moreover, since our method only fine-tunes the visual components of MLLM, it offers significant efficiency advantages. Extensive experiments demonstrate that our approach outperforms state-of-the-art unlearning methods in terms of both effectiveness and efficiency. Furthermore, we are the first to evaluate the \emph{robustness} of MLLM unlearning against re-learning attacks.
\end{abstract}

\section{Introduction}
\label{sec:intro}
Recently, Multimodal Large Language Models (MLLMs) have achieved remarkable progress. 
However, growing concerns over data privacy have become a significant barrier to the widespread application of LLMs and MLLMs. This is because training large models typically involves the use of vast amounts of Internet data, which often contain sensitive personal information such as social security numbers or personal photographs. Moreover, numerous studies~\cite{pi2024mllm,li2024digger,cohen2024performance} have shown that large models can memorize portions of their training data and reproduce them verbatim in their outputs. This poses a substantial risk of privacy leakage. Therefore, the advent of data privacy and protection regulations—such as the European Union’s GDPR~\cite{regulation2018general} and the California Consumer Privacy Act (CCPA)~\cite{goldman2020introduction}—now mandates these large model providers to honor data deletion requests from individuals~\cite{dang2021right}.

The most straightforward approach to protecting privacy is to discard the trained model entirely, remove the individual’s data from the training set, and retrain a new model from scratch. However, retraining is computationally expensive and resource-intensive, particularly for large models. As a result, the field of machine unlearning \cite{garg2020formalizing,ginart2019making,cao2015towards,gupta2021adaptive,sekhari2021remember,brophy2021machine,ullah2021machine} has emerged, 
aiming to efficiently revise trained models by selectively forgetting specific data while preserving performance on the remaining data.
Nowadays, many LLM unlearning methods have been proposed, such as Gradient Ascent (GA)~\cite{thudi2022unrolling}, its variations~\cite{liu2022continual}, and Negative Preference Optimization (NPO)~\cite{zhang2024npo}, etc. However, MLLM unlearning remains at an early stage. 

LLM unlearning problem is clearly defined—erasing the knowledge of a specific entity from the model. In contrast, defining unlearning for MLLMs is far more complex, as two distinct forms of knowledge are associated with the entity of interest: \emph{textual knowledge}, linked to the entity’s general information, and \emph{visual knowledge/visual patterns}, linked to the entity’s appearance. For instance, for a well-known individual, the name (e.g., ``Robin Williams”) and textual attributes (such as occupation or home address) constitute the textual knowledge, while the person’s facial appearance represents the visual knowledge. Consequently, there remains ambiguity in defining MLLM unlearning: some studies argue that it should involve erasing both textual and visual knowledge~\cite{liu2025MANU}, whereas others contend that only the visual knowledge should be removed while preserving the textual knowledge~\cite{huo2025mmunlearner}. 

In this paper, we follow~\cite{huo2025mmunlearner} to define the MLLM unlearning objective as \emph{erasing visual knowledge while preserving textual knowledge}, as this formulation offers greater flexibility. The essence of this definition lies in disentangling visual and textual knowledge, thereby allowing selective removal of either component when necessary. Even when the goal is to erase both textual and visual knowledge, this can be achieved by sequentially performing MLLM unlearning followed by LLM unlearning.

An MLLM typically consists of three components: a vision encoder, an LLM backbone, and a projector that bridges the two. Unlike~\cite{huo2025mmunlearner}, which fine-tunes all three components for unlearning, we propose to \emph{fine-tune only the vision encoder and the projector while keeping the LLM backbone frozen}. This design is motivated by some studies that explore the internal mechanisms of MLLMs~\cite{cohen2024performance, huang2024commonsenseknowledgeeditingbased, yu2024understanding}. These works decompose the VQA process into two stages: the first involves visually recognizing the entity (termed \emph{identification}), and the second connects the recognized entity to its stored factual knowledge (termed \emph{extraction}). It has been shown that, in a well-trained MLLM, the vision encoder and projector primarily handle the identification function, whereas the LLM module is responsible for the extraction function. Therefore, fine-tuning the visual module and disrupting the identification stage is sufficient to erase visual knowledge.
Moreover, freezing the LLM component can largely preserve textual knowledge.

Note that ~\cite{huo2025mmunlearner} conducted an ablation study comparing such \emph{partial} fine-tuning scheme (fine-tuning only the vision encoder and projector) with the \emph{full} fine-tuning scheme. Their results demonstrated that the former is inferior to the latter in balancing unlearning effectiveness and model utility. In this paper, we propose a novel approach, termed \emph{Visual Knowledge Distillation} (VKD), to harness the advantages of partial fine-tuning scheme.
Specifically, when designing the unlearning objective function, most existing methods focus primarily on the model’s output. Take GA as an example: for the forgetting knowledge, it encourages the model to avoid producing the ground-truth answer, while for the remaining knowledge, it encourages the model to retain correct predictions. We argue that such output-level supervision is well-suited for full fine-tuning scheme. However, in the partial fine-tuning scenario, this supervision signal must propagate through the frozen LLM, and thus becomes significantly weakened by the time it reaches the visual module.

To address this issue, our VKD approach injects the supervision signal directly into the visual module. Concretely, we employ the original MLLM as the teacher model and use its intermediate visual representation (\emph{i.e.}, the output of the projector) as the supervision signal. We then introduce a loss term that encourages the unlearned model (the student model) to align with this signal. Our VKD distills the teacher model’s intermediate visual representations—which predominantly encode visual knowledge—into the student model during unlearning. The empirical results demonstrate that our approach outperforms previous methods in terms of both unlearning effectiveness and model utility.

In addition to its superior unlearning performance, our partial fine-tuning scheme also offers notable advantages in efficiency. Since the knowledge distillation is applied solely to the visual component, the fine-tuning process is highly efficient, requiring updates to only a small portion of parameters. Furthermore, due to its lightweight nature, our approach is well-suited for \emph{continual} unlearning scenarios, where data deletion requests occur frequently and arrive sequentially. In such cases, our method can efficiently fine-tune the visual component on demand for each request.

Beyond evaluating the effectiveness of unlearning, we also examine its \emph{robustness}. Recent studies have revealed that machine unlearning is vulnerable to the so-called \emph{relearning attack}, which attempts to recover forgotten data or knowledge using only a few samples from the forgetting set~\cite{lynch2024robustunlearning,hu2025joggingmemory}. If the `already-unlearned' knowledge can be rapidly recovered, it suggests that the unlearning process merely obfuscates the knowledge rather than truly removes it.
Notably, this is the first work to consider the robustness of unlearning in MLLMs.

We conduct extensive experiments on representative MLLMs, including LLaVA and Qwen-VL, and compare our method with existing MLLM unlearning approaches such as MMUnlearner and MANU. Our approach surpasses these state-of-the-art methods, and interestingly, the proposed VKD module facilitates both the forgetting of target visual knowledge and the retention of non-target visual knowledge. Our method effectively disentangles textual knowledge from visual knowledge.
In addition, we evaluate the robustness of our approach against relearning attacks, demonstrating that the forgotten visual knowledge cannot be easily recovered, thereby confirming that our MLLM unlearning method possesses substantial robustness.


\section{Related Work}
\paragraph{Machine Unlearning for LLMs.}
Since LLMs may unintentionally memorize their pretraining data, Machine Unlearning (MU) techniques have been introduced to erase privacy-sensitive or copyrighted information from trained models. Gradient Ascent (GA)~\cite{thudi2022unrolling} is a representative method that fine-tunes the model to reduce the probability of generating sensitive information. One inherent challenge in unlearning lies in balancing unlearning effectiveness and model utility—that is, erasing the target data while preserving the remaining data. To address this, several GA variants have been proposed to mitigate the impact on remaining data by incorporating KL-divergence minimization~\cite{maini2024tofu} or by explicitly reducing the loss on the retained data~\cite{rafailov2023direct}.
Beyond supervised fine-tuning, several reinforcement learning-based methods, such as Negative Preference Optimization (NPO)~\cite{zhang2024npo}, have been proposed to further enhance the effectiveness of unlearning.

\paragraph{Machine Unlearning for MLLMs.}
Compared to MU for LLMs, research on MU for MLLMs remains in its infancy.
SIU~\cite{li2024single} explores the erasure of visual patterns in MLLMs using real-world entity datasets through multifaceted fine-tuning. As noted by~\cite{yu2024understanding,cohen2024performance}, MLLMs exhibit strong cross-modal interactions that entangle visual and textual knowledge, making selective unlearning more challenging and often resulting in substantial unintended knowledge loss. Moreover, even the definition of MLLM unlearning remains controversial: some studies define it as erasing visual knowledge while preserving textual knowledge~\cite{huo2025mmunlearner}, whereas others argue it should involve forgetting both visual and textual knowledge~\cite{liu2025MANU}.
Nevertheless, existing methods predominantly rely on model outputs as the supervision signal for unlearning. In contrast, our approach leverages intermediate representations through a visual knowledge distillation strategy, providing more direct and effective supervision.


\begin{figure*}[!t]
    \centering
    \includegraphics[width=5.5in]{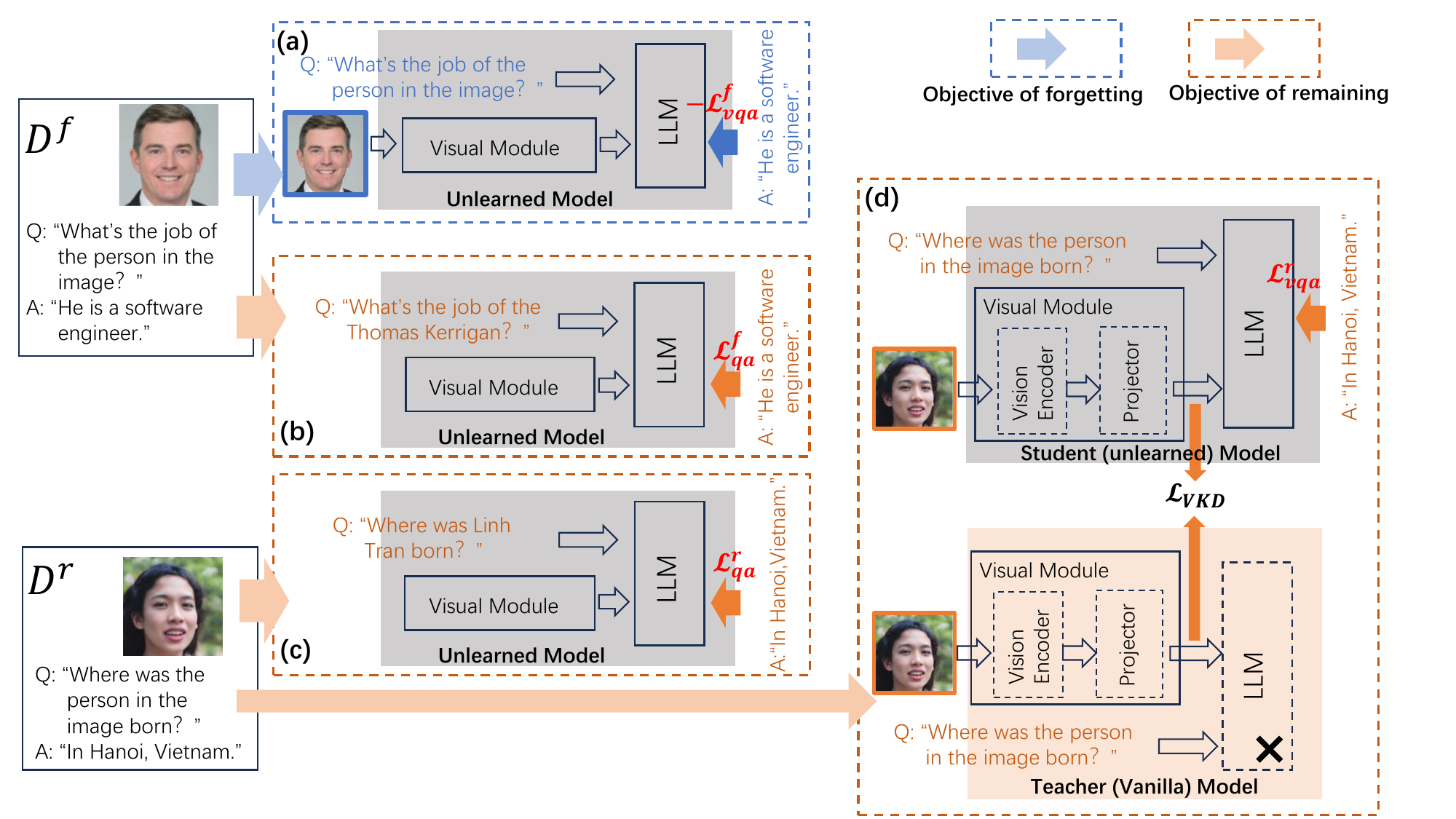}
    \caption{Illustration of our machine unlearning framework for MLLMs. For the target entity, our objective is to (a) forget its visual knowledge while (b) preserving its textual knowledge. For non-target entities, we preserve both (c) the textual knowledge and (d) the visual knowledge through VKD.}
    \label{fig1}
    \vspace{-1.0em}   
\end{figure*}

\section{Method}
\subsection{MLLM Unlearning Problem}\label{sec:def}
An MLLM model $M$ typically consists of three main components: a vision encoder $\mathcal{V}$, an LLM module $\mathcal{W}$, and a projector $\mathcal{P}$ that bridges the two. Generally, an MLLM model is trained on a visual instruction dataset $D={(v_i, q_i, a_i)}_{i=1}^N$, where $v_i$ denotes the input image, $q_i$ the textual question, and $a_i$ the corresponding answer. 

Specifically, the training of the model $M$ is carried out through a visual question answering (VQA) task as follows: (1) The image $v_i$ is first processed by the vision encoder and the projector, yielding the visual embedding $v^e_i = \mathcal{P}(\mathcal{V}(v_i))$. (2) The question $q_i$ is tokenized into a textual embedding $q^e_i$, which is then concatenated with $v^e_i$ and fed into the LLM module $\mathcal{W}(q^e_i \oplus v^e_i)$. (3) The LLM subsequently generates an output that is expected to align with the target answer $a_i$.

In the context of MLLM unlearning, we distinguish two types of knowledge associated with the entity of interest: visual knowledge and textual knowledge. We define MLLM unlearning for an entity of interest as the process of erasing its visual knowledge while preserving its textual knowledge. Additionally, both the visual and textual knowledge associated to other entities should remain unaffected. 

To evaluate the unlearning performance, we employ both VQA and QA tasks to probe the visual and textual knowledge retained by MLLMs. Specifically, given a data sample of the target entity $(v_i, q_i, a_i)$, we instruct the model to perform both the VQA and QA tasks, where the image $v_i$ is omitted in the QA task. If the model fails the VQA task (producing an incorrect answer) but succeeds in the QA task (generating the correct answer $a_i$), we regard the model as having forgotten the visual knowledge while preserving the textual knowledge of the target entity. Moreover, for the remaining data samples unrelated to the target entity, the model is expected to successfully perform both the VQA and QA tasks.

Note that conventional machine unlearning methods typically divide the training dataset $D$ into two subsets: the forgetting dataset $D^f$, which is associated with the target entity, and the remaining dataset $D^r$, which contains all other data. Thus, the unlearning problem can be formulated as:
\begin{align}
    &\mathop{\min}_{\theta} (-\mathcal{L}^f(\theta) + \mathcal{L}^r(\theta))
    \label{eq:unlearning}  
\end{align}
where $-\mathcal{L}^f(\theta)$ denotes maximizing the loss on $D^f$, $\mathcal{L}^r(\theta)$ represents minimizing the loss on $D^r$.

Clearly, this formulation is unsuitable for our MLLM unlearning objective, as the forgetting dataset $D^f$ contains both visual and textual knowledge related to the target entity. Therefore, we reformulate the unlearning objective as follows:
\begin{align}
    &\mathop{\min}_{\theta} (-\mathcal{L}^f_{vqa}(\theta) + \mathcal{L}^f_{qa}(\theta)+ \mathcal{L}^r_{vqa}(\theta)+ \mathcal{L}^r_{qa}(\theta))
    \label{eq:mllmunlearning}  
\end{align}
where $-\mathcal{L}^f_{\text{vqa}}(\theta)$ denotes maximizing the VQA loss on $D^f$, $\mathcal{L}^f_{\text{qa}}(\theta)$ represents minimizing the QA loss on $D^f$, and $\mathcal{L}^r_{\text{vqa}}(\theta)+\mathcal{L}^r_{\text{qa}}(\theta)$ represents minimizing the VQA and QA loss on $D^r$.

\subsection{Our Approach}
In this paper, we propose a novel and efficient machine unlearning algorithm to address the MLLM unlearning problem defined in the previous section. Unlike existing unlearning methods that modify the entire MLLM, our approach focuses solely on revising the visual module—that is, the vision encoder and the projector—within the MLLM. This design is motivated not only by efficiency considerations (as adjusting the visual module is significantly more lightweight than updating the entire MLLM) but also by the very definition of the MLLM unlearning objective.

Since modern MLLMs are typically built upon a pretrained LLM backbone, it is widely assumed that textual knowledge is primarily encoded in the LLM, while visual knowledge is largely captured by the visual module~\cite{zhu2023minigpt,chen2024internvl}. Moreover, one of our core unlearning goals is to preserve textual knowledge while removing visual knowledge. Therefore, freezing the parameters of the LLM naturally benefits the retention of textual knowledge.

However, when only the visual module is fine-tuned during unlearning, maintaining the delicate balance between unlearning effectiveness and model utility becomes substantially more challenging. As is well known, this trade-off has long been a central problem in machine unlearning. To mitigate it, prior studies have explored strategies such as incorporating KL-divergence regularization to preserve remaining knowledge or applying masking mechanisms to reduce conflicts between forgotten and retained knowledge. We adopt the masking mechanism in our framework; nonetheless, empirical results reveal that it alone is insufficient to meet our requirements. To this end, we introduce a new mechanism—Visual Knowledge Distillation (VKD)—to better balance unlearning and utility preservation.

\paragraph{Visual Knowledge Distillation.}
It is important to note that all existing unlearning methods generally tackle this balance at the loss level: they maximize the loss on the forgetting dataset $D^f$ (to erase target knowledge) while minimizing the loss on the remaining dataset $D^r$ (to preserve other knowledge). This paradigm is suitable when the  knowledge to be forgotten is stored across the entire model, as the \emph{output-level supervision} can backpropagate through all model components. 
However, it becomes less effective in our setting, where the visual knowledge to be erased is concentrated within the visual module, and consequently, only the visual module is fine-tuned during unlearning. Since an MLLM is structured in a cascaded manner—with the visual module feeding into the LLM backbone, the output-level supervision must propagate through the frozen LLM, and thus becomes significantly weakened by the time it reaches the visual module.

In this paper, we propose injecting supervision signals directly into the visual module, since it is the only component being updated during unlearning. We argue that such \emph{intermediate-level supervision} is more efficient and effective than conventional output-level supervision. Specifically, we use the vanilla MLLM as the teacher model, extracting its intermediate features from the visual module as supervision signals. The target model undergoing unlearning serves as the student model, which aligns its visual features with the teacher’s supervision signals for the visual knowledge to be preserved (\emph{i.e.,} $D^r$). This process, illustrated in Fig.~\ref{fig1}(d), enables the model to maintain the general visual representation capability of the teacher while selectively unlearning the undesired visual knowledge. Thus, our VKD unlearning approach can be formulated as:
\begin{align}
&\mathop{\min}_{\theta} (\mathcal{L}_{output} + \beta \mathcal{L}_{\text{VKD}}) \label{eq:prob} \\  
& \mathcal{L}_{\text{VKD}} = \|f^r_{vqa} - t^r_{vqa}\|^2 \label{eq:loss_vkd} \\  
& \mathcal{L}_{output} = -\textbf{m} \odot \alpha \mathcal{L}^f_{vqa} + \mathcal{L}^f_{qa} + \mathcal{L}^r_{vqa} + \mathcal{L}^r_{qa} \label{eq:masking}
\end{align}

where $f^r_{vqa}$ indicates the student model's visual feature for VQA input data, and $t^r_{vqa}$ indicates the teacher model's visual feature for the same inputs.

Notably, the term $||f^r_{vqa} - t^r_{vqa}||^2$ encourages the unlearned model to preserve visual knowledge unrelated to the target entity. Our empirical results indicate that when only the visual module is fine-tuned, the key to balancing unlearning effectiveness and model utility lies in retaining the visual knowledge unrelated to the target entity, while the erasure of target visual knowledge can be sufficiently achieved through output-level supervision. 
Our approach involves two weighting parameters. Specifically, $\beta$ is introduced to balance the output-level objective and the VKD objective. The parameter $\alpha$ is employed to trade off among the visual knowledge to be forgotten ($\mathcal{L}^f_{vqa}$), and the visual and textual knowledge to be preserved ($\mathcal{L}^r_{vqa}$, $\mathcal{L}^f_{qa}$ and $\mathcal{L}^r_{qa}$).

Another advantage of intermediate-level supervision is that it eliminates the need to load the full reference model into memory; instead, only the visual module is required to obtain the supervision signal.

\vspace{-0.5em}  

\paragraph{Selective Forgetting.}
As shown in Eq~\ref{eq:masking}, we follow the method~\cite{huo2025mmunlearner} that leverage a \emph{masking scheme} as performing unlearning, where $\textbf{m}$ is a boolean mask that selectively updates parameters, and $\odot$ denotes the Hadamard product.
Specifically, the mask $\textbf{m}$ is referred to as the weight saliency map, as it quantifies how salient each parameter is with respect to a given dataset $D$. In addition to saliency-based masking, another selective forgetting technique is \emph{neuron pruning}~\cite{selectivepruning}, which identifies neurons associated with the knowledge to be forgotten and achieves unlearning by pruning them. 

In this paper, we integrate these two selective forgetting strategies into a unified framework. There are multiple ways to combine them, such as parallel or sequential integration. Our experiments demonstrate that the best performance is achieved \emph{when pruning is applied first, followed by masking-based fine-tuning}. Specifically, we adopt the neuron pruning strategy based on importance functions, as proposed in~\cite{selectivepruning}. We leverage the mean of absolute activation $I_{abs}$ as the importance function. Let $n$ be a neuron and denote its corresponding activations by $z$, the importance of neuron $n$ with respect to the dataset $D$ is defined as follows:
\begin{align}
I_{\text{abs}}(D, n) = \frac{1}{|D|} \sum_{i \in D} |z(i)|
\end{align}
where $i \in D$ indicates the data sample $i$ in $D$. 

Thus, the neuron importance for unlearning $I_{u}$ is derived by the ratio of $I_{abs}$ between the knowledge to be forgotten and the knowledge to be retained:
\begin{align}
    & I_{u} = \frac{I_{abs}(D^f_{vqa})}{I_{abs}(D^f_{qa})+I_{abs}(D^r_{vqa})+I_{abs}(D^r_{qa})} \label{eq:mi}
\end{align}

Consequently, we identify the neurons to be pruned, denoted as $\textbf{n}_P$, by applying a threshold to $I_{u}$:
\begin{align}
    & \textbf{n}_P = \mathbb{I} [I_{u} \geq d_I]
\end{align}
where $d_I$ is a predefined threshold. 

After selective neuron pruning, we perform masking-based fine-tuning. The key lies in computing the saliency of each model weight with respect to the knowledge to be forgotten. Previous studies~\cite{huang2024unifiedgradientbasedmachineunlearning} have shown that this saliency can be approximated using the diagonal elements of the Fisher Information Matrix (FIM) of the vanilla model with respect to dataset $D$:
\begin{align}
    & F^D_{diag} = [\Delta \mathcal{L}^D(\theta_0)]^2
\end{align}
where $\theta_0$ denotes the vanilla model. 

Thus, the following ratio R indicates which parameters are salient for forgetting the target visual knowledge while preserving other knowledge:
\begin{align}
    & R = \frac{[\Delta \mathcal{L}^f_{vqa}]^2}{[\Delta (\mathcal{L}^f_{qa}+\mathcal{L}^r_{vqa}+\mathcal{L}^r_{qa})]^2}     
\end{align}

The Fisher mask $\textbf{m}_F$ is then obtained by comparing the ratio R against a predefined threshold $d_F$:
\begin{align}
    & \textbf{m}_F = \mathbb{I} [R \geq d_F] \label{eq:fisher}
\end{align}
This mask is regarded as the final mask $\textbf{m}$ used in Eq.~\ref{eq:masking}.

Besides this pruning-then-fine-tuning scheme, other variant strategies are discussed in Appendix. 

\paragraph{Unlearning on MLP layers.}
It is worth noting that in our approach, both neuron pruning and weight masking are applied to the visual module. This is motivated by recent studies on localizing factual memory within Transformer architectures. Prior research~\cite{liu2025MANU} has shown that MLP layers serve as primary knowledge storage components in Transformer architectures~\cite{geva2021transformerfeedforwardlayerskeyvalue, meng2023locatingeditingfactualassociations}, whereas the self-attention modules mainly function as mechanisms for information propagation~\cite{elhage2021mathematical}. 

In particular, for vision encoders such as the CLIP model, it has been observed that the high-dimensional inner projections within the feed-forward layers play a crucial role in storing visual knowledge, whereas the representations derived from the self-attention layers do not~\cite{ghiasi2022visiontransformerslearnvisual}. Furthermore, it has been observed that earlier layers learn textural attributes, whereas deeper layers learn high level object features or abstract concepts~\cite{ghiasi2022visiontransformerslearnvisual}.
Therefore, in our approach, both neuron pruning and weight masking are applied to the vision encoder's MLP layers, with emphasis on the deeper layers.


\begin{table*}[t]
\centering
\scriptsize  
\setlength{\tabcolsep}{1.0pt} 
\renewcommand{\arraystretch}{0.9} 

\begin{tabular}{l|cccccc|cccccc}
\toprule
\multirow{3}{*}{\textbf{Methods}} &
\multicolumn{6}{c|}{\textbf{MLLMU\text{-}Bench}} &
\multicolumn{6}{c}{\textbf{CLEAR}} \\

& Forget VQA & Forget QA & Retain VQA & Retain QA & Realworld VQA & Realworld QA
& Forget VQA & Forget QA & Retain VQA & Retain QA & Realworld VQA & Realworld QA \\
& Acc($\downarrow$) & Acc($\uparrow$) & Acc($\uparrow$) & Acc($\uparrow$) & Acc($\uparrow$) & Acc($\uparrow$)
& Acc($\downarrow$) & R\text{-}L($\uparrow$) & Acc($\uparrow$) & R\text{-}L($\uparrow$) & Acc($\uparrow$) & Acc($\uparrow$) \\
\midrule
\multicolumn{13}{c}{\textbf{LLaVA-1.5-7B}} \\
\midrule
Vanilla      & 45.8 & 38.4 & 45.2 & 37.5 & 47.4 & 54.9 & 63.3 & 0.367 & 54.0 & 0.352 & 53.7 & 85.4 \\
GA           & 43.2 & 32.5 & \textbf{45.0} & 32.2 & \underline{47.0} & 55.0 & 57.4 & 0.153 & \underline{48.4} & 0.176 & \underline{51.8} & 83.4 \\
GA\_Diff     & 40.0 & 33.6 & 44.3 & 31.5 & 46.6 & 53.6 & 47.3 & 0.197 & 43.4 & 0.220 & 47.7 & 73.5 \\
KL\_Min      & 42.4 & 33.6 & 44.9 & 32.0 & 46.6 & 54.6 & 40.4 & 0.270 & 38.1 & 0.274 & 51.5 & 82.8 \\
NPO          & 43.2 & 33.6 & \underline{44.9} & 32.2 & \underline{47.0} & 55.0 & 40.4 & 0.285 & 38.6 & 0.282 & \underline{51.8} & 83.4 \\
MANU         & 41.6  & 33.6  & 43.2  & 32.1  & 46.3  & \underline{54.8}  & 39.8  & 0.263 & 25.8  & 0.224 & 49.8  & 68.3  \\
MMU  & \underline{31.2} & \underline{34.2} & 44.2 & \underline{35.1} & 46.7 & \textbf{54.9} & \underline{36.2} & \underline{0.348} & 46.6 & \underline{0.338} & 51.2 & \underline{84.1} \\
\rowcolor{lightgray}
Ours         & \textbf{29.8} & \textbf{35.8} & \textbf{45.0} & \textbf{36.1} & \textbf{47.2} & \textbf{54.9} & \textbf{34.2} & \textbf{0.349} & \textbf{49.6} & \textbf{0.346} & \textbf{52.8} & \textbf{84.8} \\
\midrule
\multicolumn{13}{c}{\textbf{Qwen2\text{-}VL\text{-}7B}} \\
\midrule
Vanilla      & 55.2 & 55.0 & 56.0 & 58.6 & 77.3 & 77.5 & 67.0 & 0.116 & 70.9 & 0.098 & 69.2 & 91.4 \\
GA           & 50.4 &  \underline{46.7} & \underline{51.5} & \textbf{57.6} & 74.4 & 77.8 & 55.3 & \underline{0.123} & 62.4 & 0.083 & 65.9 & 86.8 \\
GA\_Diff     & 54.4 & 52.8 & 38.8 & 54.4 & 74.5 & 77.0 & 63.3 & \textbf{0.125} & 64.4 & 0.088 & 66.0 & \underline{92.7} \\
KL\_Min      & 45.6 & 45.3 & 35.9 & 55.6 & 74.8 & 77.1 & 67.0 & 0.120 & \underline{68.9} & \underline{0.098} & 68.4 & 90.7 \\
NPO          & 49.6 & 50.4 & 49.5 & 53.3 & 75.2 & \textbf{78.3} & 62.8 & 0.103 & 68.3 & 0.091 & \underline{68.9} & 88.7 \\
MANU         & \underline{43.2}  & 47.2  & 43.2  & 49.4  & 73.7  & 76.2  & 61.3  & 0.107 & 65.2  & 0.094 & 66.2  & 87.3  \\
MMU  & 44.0 & 54.4 & \underline{52.6} & 55.7 & \underline{75.3} & 77.3 & \underline{50.0} & \underline{0.123} & 68.3 & \textbf{0.100} & \underline{68.9} & \textbf{94.7} \\
\rowcolor{lightgray}
Ours         & \textbf{42.8} & \textbf{54.6} & \textbf{56.0} & \underline{56.4} & \textbf{75.4} & \underline{77.4} & \textbf{48.7} & \textbf{0.125} & \textbf{70.9} & 0.094 & \textbf{69.0} & 90.8 \\
\bottomrule
\end{tabular}
\caption{Overall results of baselines and our approach on MLLMU\text{-}Bench and CLEAR benchmarks. \textbf{Bold} and \underline{underlined} values indicate the best and second-best performance, respectively. $\downarrow$ Lower is better for Forget VQA accuracy, $\uparrow$ Higher is better for the others. For MLLMU-Bench, the Forget/Retain QA tasks are evaluated using the Acc metric, while for CLEAR, these tasks are evaluated using the ROUGE-L metric.}
\label{tab1}
\vspace{-1.5em} 
\end{table*}

\section{Evaluation}
To accurately assess whether the target knowledge has been successfully unlearned from the MLLM, we typically do not perform unlearning directly on the original pretrained model (\emph{e.g.,} LLaVA-1.5-7B), as it is uncertain whether the model initially possesses the target knowledge. In practice, most unlearning benchmarks adopt a two-stage procedure: first, a fictitious dataset representing the target knowledge (\emph{e.g.,} synthetic personal profiles) is curated, and the \emph{base} MLLM is fine-tuned on this dataset to obtain a \emph{vanilla} model that explicitly contains the target knowledge. Then, the unlearning algorithm is applied to this vanilla model, and the resulting \emph{unlearned} model is evaluated to assess the effectiveness of the unlearning algorithm.

\subsection{Datasets and Metrics}
\textbf{Datasets:} We employ two benchmark datasets to evaluate unlearning performance in MLLMs. \textbf{MLLMU-Bench} comprises fictitious personal profiles, each paired with a portrait and 18 corresponding questions (\emph{i.e.}, 9 VQA questions and 9 textual QA questions) with multiple-choice options. The dataset is divided into three subsets: Forget, Retain, and Real-world. The Forget Set contains a subset of fictitious profiles designated for evaluating unlearning effectiveness. The Retain Set includes fictitious profiles excluded from the Forget Set, representing the knowledge the model should preserve. The Real-world Set consists of real celebrity profiles distinct from the fictitious ones and is used to assess unlearning entanglement with neighboring real-world concepts, thereby reflecting the model’s utility. \textbf{CLEAR} is built on top of an LLM unlearning dataset, \emph{i.e.,} the TOFU dataset. For each author in TOFU, CLEAR adds several face images and corresponding textual descriptions generated by GPT-4o \cite{achiam2023gpt}. Similar to MLLMU-Bench, CLEAR also comprises Forget, Retain, and Real-world sets.

\textbf{Metrics:}
We evaluate the forgetting and retention of textual knowledge using the Question Answering (QA) task, and assess the forgetting and retention of visual knowledge using the Visual Question Answering (VQA) task. The performance on multiple-choice questions is measured by \emph{average accuracy}, while text generation performance is evaluated using the \emph{ROUGE-L} metric~\cite{lin2004rouge}.

Beyond evaluating the \emph{effectiveness} of machine unlearning, we further investigate its \emph{robustness}. Prior studies have demonstrated that \emph{LLM unlearning} is susceptible to \emph{relearning attacks}, which aim to restore forgotten knowledge using only a few samples from the forgetting set. To the best of our knowledge, our work is the first to explore the \emph{robustness} of unlearning in MLLMs. Specifically, after performing MLLM unlearning, we fine-tune the unlearned model with a small subset of the forgotten visual knowledge and subsequently assess the extent to which this forgotten knowledge can be recovered through a relearning process.

\subsection{Models and Baselines}
\textbf{MLLMs:} We conduct experiments using open-source MLLMs as base models. Specifically, we use LLaVA-1.5-7B-hf and Qwen2-VL-7B-Instruct, which serve as the foundation for our unlearning experiments. The vanilla models are trained following the official implementations provided by MLLMU-Bench and CLEAR. 

\textbf{Baselines:}
We compare our method with six baseline approaches: 
\textbf{GA~\cite{thudi2022unrolling}}, which applies gradient ascent on the forget VQA set $D_f$;
\textbf{GA\_Diff~\cite{liu2022continual}}, an improved GA variant with a joint loss to balance the forget set $D_f$ and retain set $D_r$;
\textbf{KL\_Min~\cite{maini2024tofu}}, which aligns predictions on $D_r$ with the original model while maximizing KL divergence with $D_f$;
\textbf{NPO~\cite{zhang2024npo}}, which treats $D_f$ as negative-preference data and integrates unlearning into a preference optimization framework with an oracle model fine-tuned on $D_r$;
\textbf{MANU~\cite{liu2025MANU}}, which introduces modality-aware neuron unlearning by selectively pruning neurons that contribute most to the forget set, preserving model performance while achieving balanced forgetting across modalities;
\textbf{MMU(nlearner)~\cite{huo2025mmunlearner}}, which uses saliency-weighted updates to selectively protect non-target parameters while forgetting visual knowledge. 


\begin{figure}[!t]
    \centering
    \includegraphics[width=3.0in]{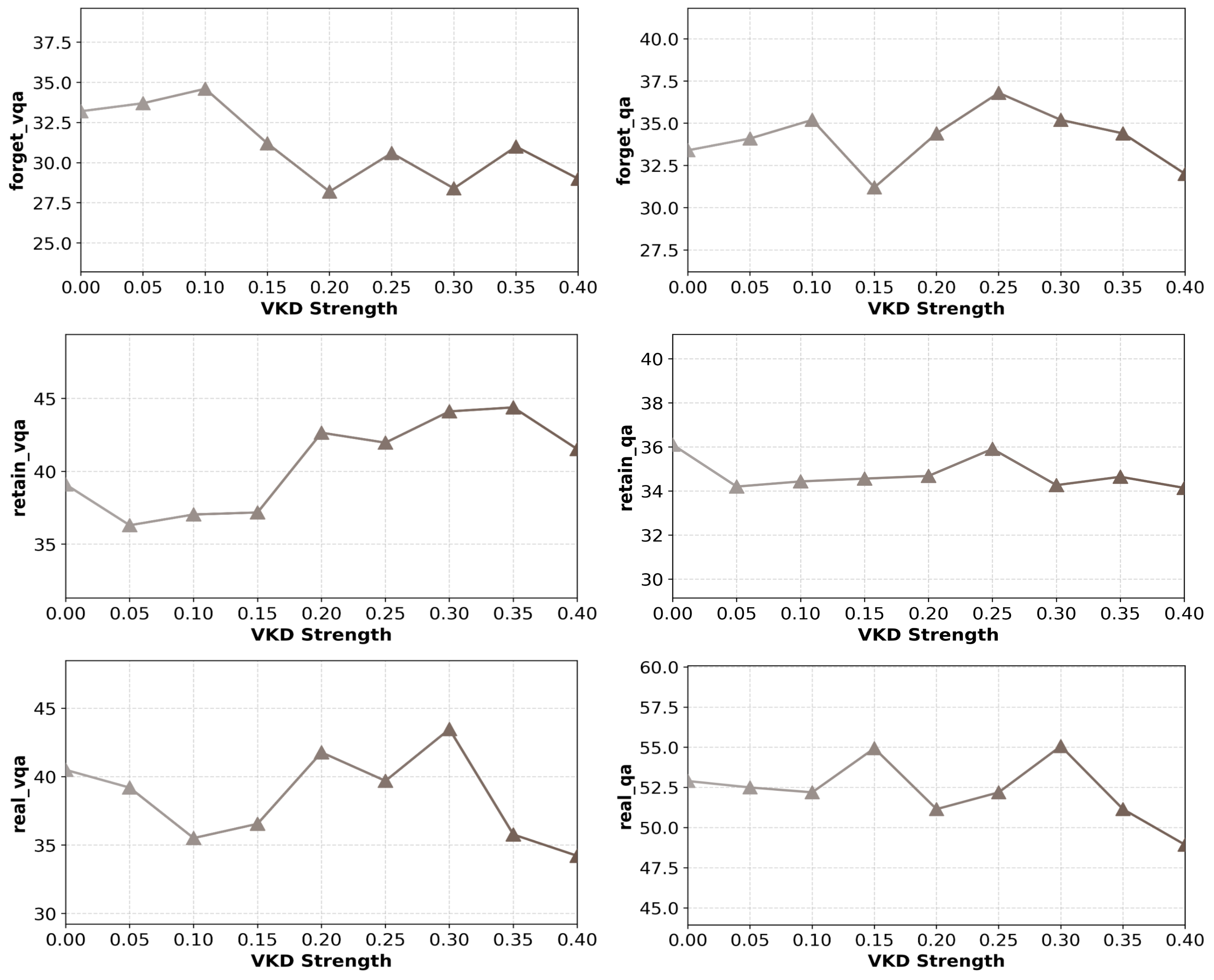}
    \caption{The trade-off between the output-level objective and the VKD objective with respect to the VKD strength $\beta$.}
    \label{fig2}
\vspace{-1.0em}    
\end{figure}

\subsection{Main Results and Analysis}

\paragraph{Effectiveness of Unlearning.}
As shown in Table~\ref{tab1}, our approach can significantly reduce the accuracy of Forget VQA task on MLLMU-Bench for both LLaVA-7B and Qwen2-VL, surpassing the state-of-the-art method MMUnlearner by \textbf{1.4\%} and \textbf{1.2\%}, respectively. Similarly, on CLEAR, our approach consistently surpasses MMUnlearner by \textbf{2.0\%} and \textbf{1.3\%}, respectively. This illustrates that our approach can effectively erase the visual knowledge associated with the target entity.

Our method also retains the accuracy of both the Retain QA task and the Forget QA task, particularly in MLLMU-Bench dataset. This demonstrates that our approach effectively preserves the textual knowledge associated with both target and non-target entities.

Compared to target visual knowledge forgetting, retaining the visual knowledge associated with non-target entities remains the most challenging issue. From the results on the Retain VQA and Real-world VQA, our approach shows substantial improvement, outperforming both MANU and MMUnlearner. This demonstrates that our method possesses a strong advantage in balancing unlearning effectiveness and model utility with respect to visual knowledge.

We present several examples for comparing our approach with GA, NPO, and MANU, as shown in Appendix. We find that our method can successfully erases the target visual content. In contrast, GA often leads to partial forgetting and distorted textual responses, whereas NPO tends to over-forget, degrading both visual and textual modalities.

\begin{figure}[!t]
    \centering
    \includegraphics[width=3.0in]{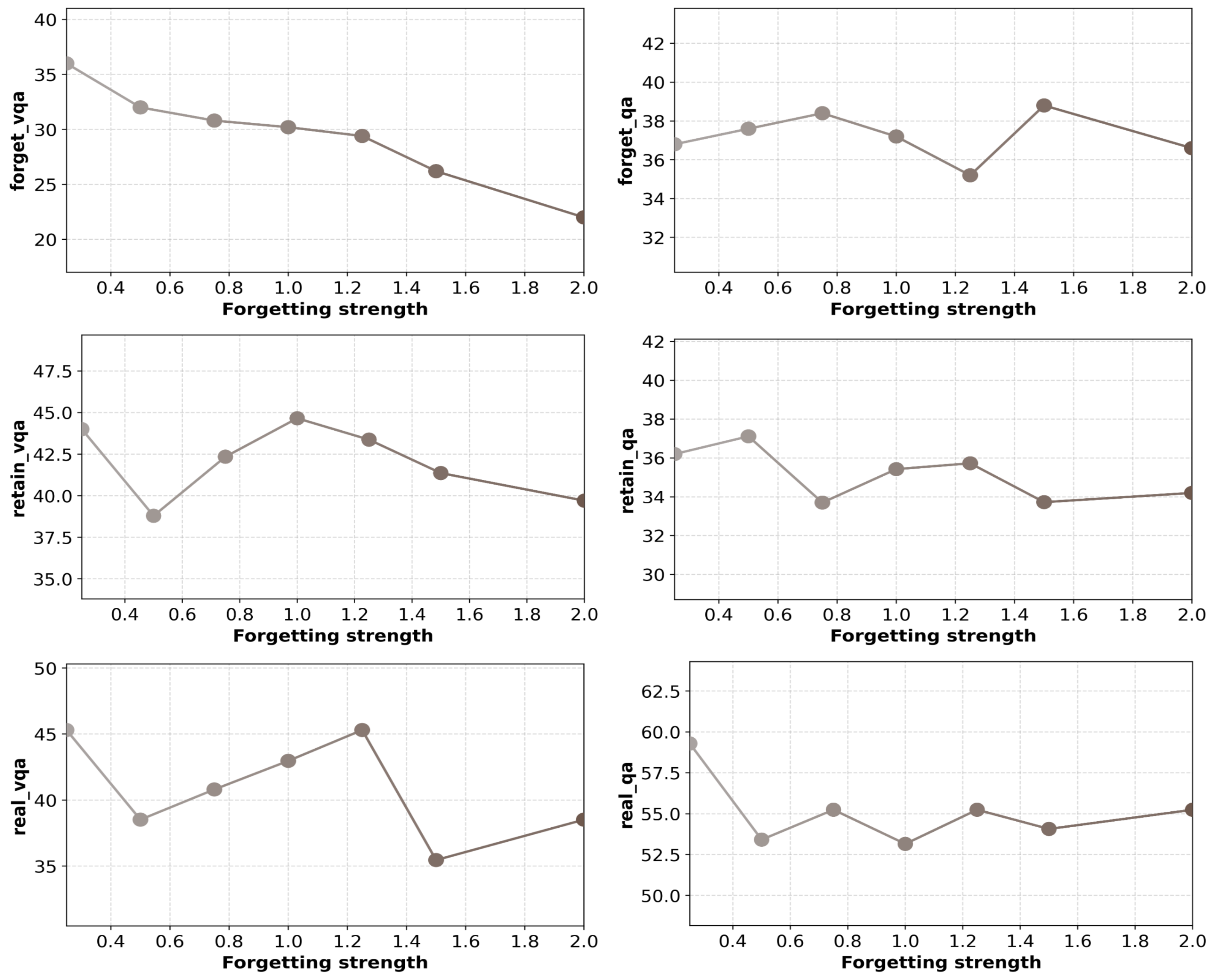}
    \caption{The trade-off between knowledge forgetting and preservation with respect to the weighting parameter $\alpha$.}
    \label{fig4}
\end{figure}

\paragraph{Effect of VKD.}
The VKD module is designed to preserve the visual knowledge associated with non-target entities. Its strength is controlled by the parameter $\beta$ in Eq.~\ref{eq:prob}. As shown in Fig.~\ref{fig2}, when $\beta=0$—that is, when the VKD term in Eq.~\ref{eq:prob} is omitted—the visual knowledge related to non-target entities is not effectively preserved, as evidenced by the notably low accuracy on the Retain VQA task. In contrast, as $\beta$ increases, the accuracy of the Retain VQA rises sharply, indicating a substantial improvement in preserving the visual knowledge associated with non-target entities.

Interestingly, \emph{our VKD module also contributes to forgetting the visual knowledge associated with the target entity, even though it was not originally designed for this purpose}. As shown in Fig.~\ref{fig2}, as $\beta$ increases, the accuracy of the Forget VQA also exhibits a noticeable decline. These results illustrate that our approach effectively disentangles the forgetting of target visual knowledge from the preservation of non-target visual knowledge. 

Notably, the strength of VKD should not be set excessively high. We observe that as $\beta$ increases, the accuracy of the Real-world VQA initially rises but subsequently declines, indicating that an overly strong VKD may harm the model’s overall utility. Finally, we set the optimal strength to $\beta = 0.3$, which achieves a balanced trade-off between the output-level objective and the VKD objective.

Moreover, based on the accuracy of the Forget QA, Retain QA, and Real-world QA, we observe that they are only slightly influenced by the VKD module. This demonstrates that our VKD design does not compromise the preservation of textual knowledge.
\vspace{-1.2em}

\paragraph{Balance between Forgetting and Preservation.}
One inherent challenge in machine unlearning lies in balancing knowledge forgetting and preservation. Our approach employs two complementary strategies to address this issue. The first strategy involves selectively fine-tuning only a subset of parameters during unlearning, controlled by the mask $\textbf{m}$ in Eq.~\ref{eq:masking}. The second strategy introduces the weighting parameter $\alpha$ in Eq.~\ref{eq:masking} to regulate the relative strengths of forgetting and preservation objectives. We analyze the second strategy here and will discuss the first strategy in the following section.

\begin{table}[tpb]
\centering
\scriptsize
\setlength{\tabcolsep}{1.2pt}
\renewcommand{\arraystretch}{0.9}
\begin{tabular}{lcccc}
\toprule
\textbf{Methods} & \textbf{Fine-tuning Scope} & \textbf{Running Time} & \textbf{GPU memory}\\
\midrule
GA           & Full Model (Vision + LLM)   & 43.3~($\pm$1.5)min & 48.7~($\pm$0.38)GB \\
GA\_Diff      & Full Model (Vision + LLM)   & 57.4~($\pm$1.5)min & 50.1~($\pm$0.37)GB  \\
NPO           & Full Model (Vision + LLM)   & 61.3~($\pm$1.5)min & 68.3~($\pm$0.38)GB  \\
MMU          & Full Model (Vision + LLM)    & 23.6~($\pm$1.5)min & 50.5~($\pm$0.38)GB\\
\midrule
\rowcolor{lightgray}
Ours (VKD)           & Vision & 16.7~($\pm$1.3)min & 47.7~($\pm$0.24)GB\\
\rowcolor{lightgray}
Ours (w/o VKD)       & Vision    & \textbf{15.4~($\pm$1.3)min} & \textbf{46.7~($\pm$0.24)GB}\\
\bottomrule
\end{tabular}
\caption{Comparison of Unlearning Efficiency. Average running time per epoch (in minutes) on NVIDIA RTX 4090 GPU. The GPU memory consumption is compared across these methods.}
\label{tab4}
\vspace{-2.5em}  
\end{table}

The parameter $\alpha$ controls the strength of target visual knowledge forgetting relative to the preservation of other forms of knowledge, including the target textual knowledge (corresponding to $\mathcal{L}^f_{qa}$) and both the visual and textual knowledge associated with non-target entities (corresponding to $\mathcal{L}^r_{vqa}$ and $\mathcal{L}^r_{qa}$). As shown in Fig.~\ref{fig4}, as $\alpha$ increases, the accuracy of the Forget VQA drops rapidly, indicating that the forgetting of target visual knowledge is significantly enhanced. Meanwhile, the accuracies of the Retain VQA and Real-world VQA initially rise but subsequently decline, reflecting the inherent conflict between knowledge forgetting and preservation. Nevertheless, a proper forgetting strength of $\alpha = 1.25$ achieves a desirable balance between the two objectives.

As shown in Fig.~\ref{fig4}, the accuracies of the Forget QA, Retain QA, and Real-world QA remain largely insensitive to variations in $\alpha$, indicating that our approach effectively disentangles textual knowledge from visual knowledge.

\paragraph{Efficiency of Unlearning.}
By freezing the LLM backbone and applying lightweight fine-tuning to the visual module, our method achieves notable efficiency gains.
We measure the average per-epoch unlearning time under different configurations and compare with other baseline methods on NVIDIA 4090 GPU.
As shown in Table~\ref{tab4}, our approach achieves a favorable trade-off between unlearning efficacy and computational efficiency.
Freezing the LLM \emph{reduces the overall unlearning time by more than 50\%}, while maintaining strong unlearning effectiveness. The introduction of VKD adds only minor overhead.

\begin{table}[tpb]
\centering
\scriptsize  
\setlength{\tabcolsep}{1.0pt} 
\renewcommand{\arraystretch}{0.80} 
\begin{tabular}{l|cc|cc|cc}
\toprule
\multirow{3}{*}{\textbf{Methods}} &
\multicolumn{2}{c|}{\textbf{Forget Set}} &
\multicolumn{2}{c|}{\textbf{Retain Set}} &
\multicolumn{2}{c}{\textbf{Realworld Set}} \\
& Forget VQA & Forget QA & Retain VQA & Retain QA & RW VQA & RW QA \\
& Acc($\downarrow$) & Acc($\uparrow$) & Acc($\uparrow$) & Acc($\uparrow$) & Acc($\uparrow$) & Acc($\uparrow$) \\
\midrule
Vanilla                     & 45.8 & 38.4 & 45.2 & 37.5 & 47.4 & 54.9 \\
B & 40.0 & 33.6 & 44.3 & 31.5 & 46.6 & 53.6 \\
F                  & 15.8 & 24.0 & 20.5 & 35.0 & 19.8 & 45.6 \\
B+Mask                        & 36.4 & 32.7 & 43.6 & 36.7 & 37.8 & 50.6 \\
F+Mask              & 33.2 & 33.4 & 39.1 & 36.1 & 40.5 & 52.9 \\
F+VKD      & 30.8 & 30.6 & 40.7 & 34.9 & 44.8 & 53.5 \\
B+Mask+VKD          & 35.6 & 31.8 & 42.1 & 34.1 & 45.9 & 52.6 \\
\rowcolor{lightgray}
F+Mask+VKD & 29.8 & 35.8 & 45.0 & 36.1 & 46.9 & 54.9 \\
\bottomrule
\end{tabular}
\caption{Ablation study. Lower is better ($\downarrow$) for Forget VQA accuracy; higher is better ($\uparrow$) for the others. “B” denotes the baseline, which fine-tunes the entire MLLM model, whereas “F” denotes freezing the LLM while fine-tuning only the visual module. “F+mask+VKD” represents our proposed approach.}
\label{tab2}
\vspace{-0.5em} 
\end{table}

\begin{table}[t]
\centering
\scriptsize
\setlength{\tabcolsep}{2pt} 
\renewcommand{\arraystretch}{0.9} 
\begin{tabular}{llccccccc}
\toprule
\textbf{Task} & \textbf{Method} & \textbf{Original} & \textbf{Epoch1} & \textbf{Epoch2} & \textbf{Epoch3} & \textbf{Epoch4} & \textbf{Epoch5} & \textbf{AG}\\
\midrule
\multirow{3}{*}{Forget VQA} 
 & MMU & 29.1& 31.2& 35.4 & 37.4 & 36.2  & 37.4  & 8.3
\\
 & MANU &38.6 &38.2 &40.8  &44.6 &42.9 &42.9 & 4.3
\\
 & Ours & 29.3 &29.2  &27.6  & 27.6  &30.6  &30.6  & \textbf{1.3}
\\
\midrule
\multirow{3}{*}{Forget QA} 
 & MMU & 43.6  & 47.6  & 54.4  & 52.0 & 53.4  & 53.4  & 9.8
\\
 & MANU  & 31.5 &35.6  & 33.6  & 36.4  & 36.4  & 37.6  & 6.1
\\
 & Ours  & 43.2 & 43.6  & 42.8  & 43.2  & 44.0 & 44.0 & \textbf{0.8}\\
\bottomrule
\end{tabular}
\caption{Comparison of unlearning robustness for different methods over 5 epochs relearning attack.}
\label{tab:forget_epochs}
\vspace{-2.5em}
\end{table}

\subsection{Ablation Study}
Our approach incorporates several strategies to enhance MLLM unlearning, including partial fine-tuning, visual knowledge distillation, and the saliency masking scheme. We conduct an ablation study to demonstrate the contribution of each component and their complementary effects.

Unlike previous methods that fine-tune the entire MLLM or all MLP components during unlearning, we adopt a partial fine-tuning strategy: only the vision module—specifically, the MLP layers within the vision module—is fine-tuned, while the LLM backbone parameters remain frozen. We argue that this strategy offers efficiency advantages for unlearning, as the vision module is significantly smaller than the LLM backbone. Moreover, keeping the LLM frozen helps preserve the textual knowledge.

As shown in Table~\ref{tab2}, the baseline method (\emph{i.e.,} the “B” row) adopts a full fine-tuning strategy, wherein both the visual module and the LLM backbone are fine-tuned. When partial fine-tuning is excluded and only the VKD and masking schemes are applied (\emph{i.e.,} the “B+Mask+VKD” row), we observe that this configuration is inferior to our approach (\emph{i.e.,} the “F+Mask+VKD” row) in preserving textual knowledge, as evidenced by the lower accuracy on the Forget QA task. Moreover, its running time is significantly longer than that of our approach.

Comparing “F+Mask” to “F+VKD,” we observe that the masking scheme excels at preserving textual knowledge for both target and non-target entities, as reflected by the accuracy on the Forget and Retain QA sets. In contrast, “F+VKD” demonstrates superior performance in preserving visual knowledge associated with non-target entities and forgetting visual knowledge associated with the target entity. Thus, \emph{the masking and VKD strategies complement each other}, enabling our approach (\emph{i.e.,} “F+Mask+VKD”) to outperform either strategy used individually.

We integrate two complementary schemes for selective forgetting: neuron pruning and weight masking.
We conduct an ablation study by applying each scheme independently. The results demonstrate that our integration-based strategy outperforms any single scheme. Details of the ablation study are provided in Appendix.

\subsection{Robustness of Unlearning}
When conducting a relearning attack using a small portion of the forgetting data, we evaluate the extent to which the forgotten visual knowledge can be recovered after the attack. Specifically, we define a new metric, \emph{Accuracy Gap} (AG), defined as the difference between the \emph{post-relearning accuracy} and the accuracy of the original model, where the post-relearning accuracy refers to the final accuracy once relearning procedure has stably converged. The smaller the AG, the more robust the unlearning algorithm is. In practice, we measure the AG using the accuracy on the Forget VQA task, \emph{i.e.}, by comparing the difference in Forget VQA accuracy before and after the relearning process.

We conduct relearning attacks using 10\%, 20\%, and 30\% of the forgotten visual data. As shown in Table~\ref{tab:forget_epochs}, as the proportion of fine-tuning data used for the relearning attack increases, more forgotten knowledge is gradually recovered. However, with 20\% of the relearning data, the Forget VQA accuracy shows only a slight recovery, increasing from 29.3\% to 30.6\% (\emph{i.e.,} AG=1.3\%), demonstrating that our approach effectively forgets the undesired knowledge and remains robust against relearning attacks. More results are provided in Appendix C.

\section{Conclusion}
In this paper, we propose an MLLM unlearning approach that fine-tunes only the visual module while keeping the LLM backbone frozen.
We introduce a VKD scheme, which incorporates intermediate visual representation from a teacher model as a supervision signal. Interestingly, VKD not only facilitates forgetting of target visual knowledge but also aids in preserving visual knowledge of non-target entities. Meanwhile, we combine neuron pruning and FIM masking in a unified framework. 
Moreover, we evaluate the robustness of our MLLM unlearning approach against relearning attacks, demonstrating that the forgotten visual knowledge cannot be easily recovered.

\bibliography{example_paper}
\bibliographystyle{icml2026}

\end{document}